\documentclass[letterpaper, 10 pt, conference]{ieeeconf}  

\IEEEoverridecommandlockouts                              
\usepackage{xcolor}
\usepackage[normalem]{ulem}

\title{\LARGE \bf
Improved Exploration through Latent Trajectory Optimization in Deep Deterministic Policy Gradient
}

\author{Kevin Sebastian Luck$^{1}$, Mel Vecerik$^{2}$, Simon Stepputtis$^{1}$, Heni Ben Amor$^{1}$ and Jonathan Scholz$^{2}$%
\thanks{$^{1}$Interactive Robotics Lab, Arizona State University, Tempe, AZ, USA \newline
      {\tt\small $\lbrace \text{ksluck},~\text{sstepput},~\text{hbenamor}\rbrace$ \"at asu.edu} 
      }%
\thanks{$^{2}$Google DeepMind, London, UK. \newline
      {\tt\small $\lbrace \text{vec},~\text{jscholz} \rbrace$ \"at google.com}}%
}

\usepackage{amsmath}
\usepackage{amsfonts}
\usepackage{todonotes}
\usepackage{graphicx}
\usepackage{subcaption}
\usepackage{algorithmic}
\usepackage{algorithm}

\newcommand{\klvec}[1]{\ensuremath{\mathbf{#1}}}
\newcommand{\klim}{\ensuremath{\mathbf{\text{Im}}}}
\newcommand{\klenc}{\ensuremath{E}}
\newcommand{\kldec}{\ensuremath{\text{D}}}
\newcommand{\kldyn}{\ensuremath{\Psi}}
\newcommand{\kllat}{\ensuremath{\mathbf{z}}}
\newcommand{\klaction}{\ensuremath{\mathbf{a}}}

\newcommand{\klcriticV}{\ensuremath{\text{V}}}
\newcommand{\klcriticQ}{\ensuremath{\text{Q}}}
\newcommand{\klact}{\ensuremath{\pi}}

\begin{document}

\maketitle
\thispagestyle{empty}
\pagestyle{empty}

\begin{abstract}
Model-free reinforcement learning algorithms such as Deep Deterministic Policy Gradient (DDPG) often require additional exploration strategies, especially if the actor is of deterministic nature. This work evaluates the use of model-based trajectory optimization methods used for exploration in Deep Deterministic Policy Gradient when trained on a latent image embedding. In addition, an extension of DDPG is derived using a value function as critic, making use of a learned deep dynamics model to compute the policy gradient. This approach leads to a symbiotic relationship between the deep reinforcement learning algorithm and the latent trajectory optimizer. The trajectory optimizer benefits from the critic learned by the RL algorithm and the latter from the enhanced exploration generated by the planner. The developed methods are evaluated on two continuous control tasks, one in simulation and one in the real world. In particular, a Baxter robot is trained to perform an insertion task, while only receiving sparse rewards and images as observations from the environment.
\end{abstract}


\section{Introduction}
Reinforcement learning (RL) methods enabled the development of autonomous systems that can autonomously learn and master a task when provided with an objective function. 
RL has been successfully applied to a wide range of tasks including flying \cite{tedrake2009learning, reddy2018glider}, manipulation \cite{vevcerik2017leveraging, li2019robot,  luck2017extracting, colome2019exploiting, chebotar2018closing}, locomotion \cite{li2018using, luck2017lab2desert}, and even autonomous driving~\cite{jaritz2018end, kendall2018learning}.
The vast majority of RL algorithms can be classified into the two categories of (a) inherently stochastic or (b) deterministic methods. 
While inherently stochastic methods have their exploration typically built-in~\cite{haarnoja2018soft, schulman2017proximal}, their deterministic counterparts require an, often independent, exploration strategy for the acquisition of new experiences within the task domain~\cite{lillicrap2015continuous,kober2009policy}. 
In deep reinforcement learning, simple exploration strategies such as Gaussian noise or Ornstein-Uhlenbeck (OU) processes \cite{uhlenbeck1930theory}, which model Brownian motion, are standard practice and have been found to be effective \cite{lillicrap2015continuous}. 
However, research has shown that advanced exploration strategies can lead to a higher sample-efficiency and performance of the underlying RL algorithm \cite{luck2014latent}. 
\begin{figure}
    \centering
    \includegraphics[width=0.42\textwidth]{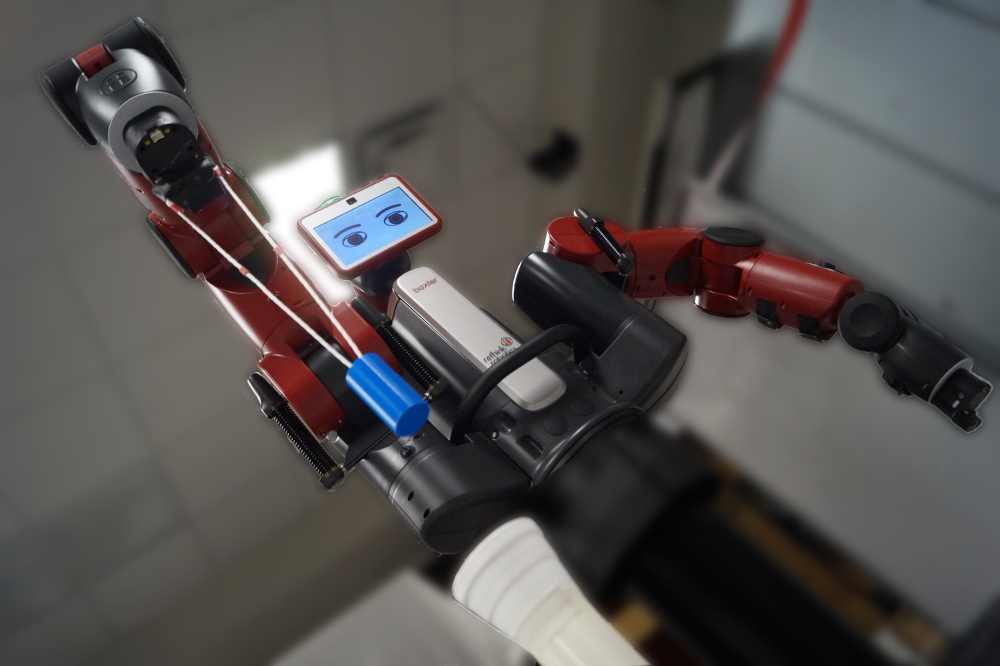}
    \caption{A Baxter robot learning a visuo-motor policy for an insertion task using efficient exploration in latent spaces. The peg is suspended from a string.
    }
    \label{Fig::teaser}
\end{figure}
In practice, there are two ways to incorporate advanced exploration strategies into deterministic policy search methods. 
Where possible, one can reformulate the deterministic approach within a stochastic framework, such as by modeling the actions to be sampled as a distribution. 
Parameters of the distribution can then be trained and are tightly interconnected with the learning framework. 
One example for this methodology, is the transformation of Policy Search with Weighted Returns (PoWER)~\cite{kober2009policy} into Policy Search with Probabilistic Principal Component Exploration (PePPEr)~\cite{luck2014latent}. 
Instead of using a fixed Gaussian distribution for exploration, the noise generating process in PePPEr is based on Probabilistic Principal Component Analysis (PPCA) and generates samples along the latent space of high-reward actions. 
Generating explorative noise from PPCA and sampling along the latent space was shown to outperform the previously fixed Gaussian exploration. 
Alternatively, one can choose to optimize the exploration strategy itself. 
Examples of this methodology are count-based exploration strategies~\cite{tang2017exploration}, novelty search~\cite{stadie2015incentivizing} or curiosity-driven approaches \cite{pathak2017curiosity} which can be transferred with ease to other algorithms or frameworks.
Typically, when incorporating these techniques into reinforcement learning, they are limited to local exploration cues based on the current state. 
This paper aims to combine the model-free deep deterministic policy gradient method with a model-based exploration technique for increased sample-efficiency in real world task domains.
The proposed method generates exploratory noise by optimizing a (latent) trajectory from the current state to ideal future states, based on value functions learned by an RL algorithm. 
This experience is, in turn, used by the RL algorithm to optimize policy and value functions in an off-policy fashion, providing an improved objective function for the trajectory optimizer. 
We investigate whether this strategy of formulating exploration as a latent trajectory optimization problem leads to an improved learning process both in simulation, as well as in a robotic insertion task executed solely in the real world. 
In particular, we apply our approach to a challenging, flexible insertion task as seen in Fig.~\ref{Fig::teaser}. 

\begin{figure}
    \centering
    \begin{subfigure}[b]{0.185\textwidth}
        \centering
        \includegraphics[width=0.95\textwidth]{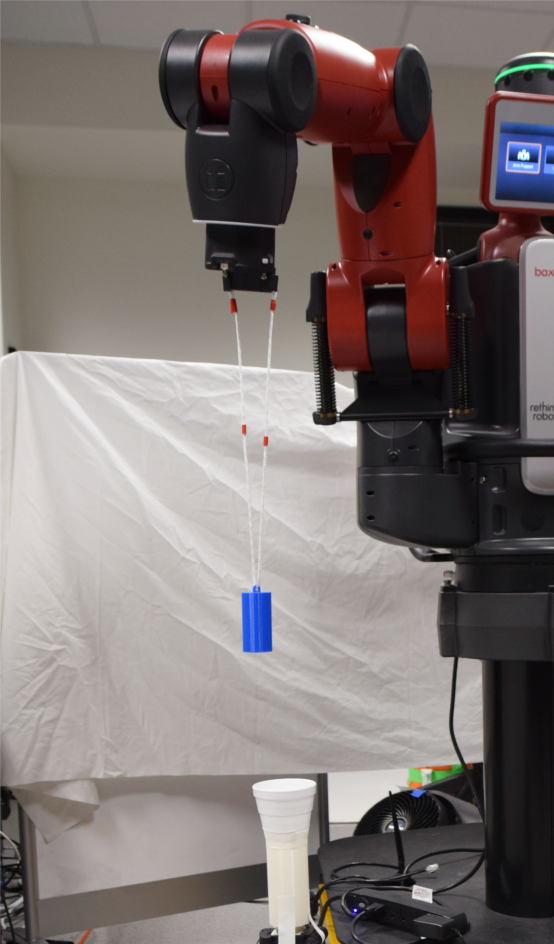}
        \caption{Rand. initial position}
    \end{subfigure}%
    \begin{subfigure}[b]{0.3032\textwidth}
        \centering
        \includegraphics[width=0.95\textwidth]{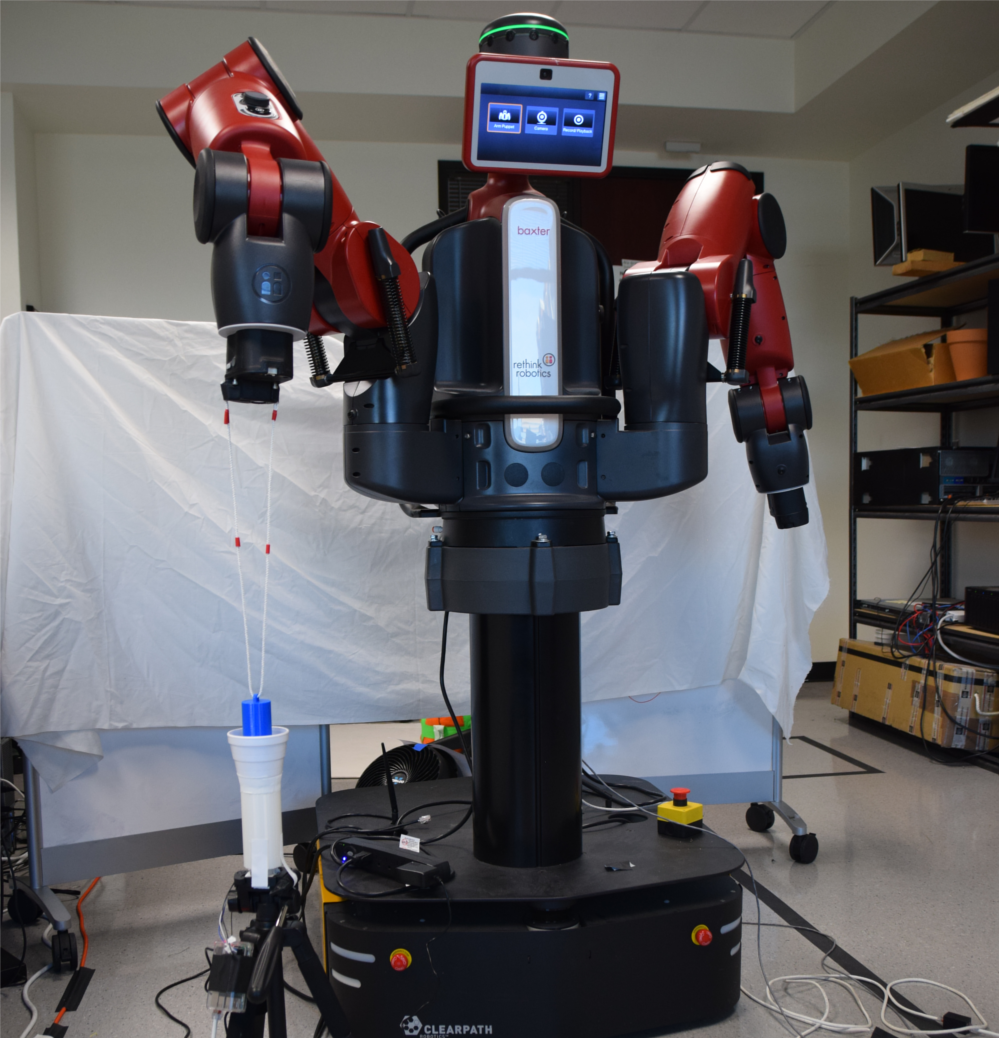}
        \caption{Insertion started}
    \end{subfigure}
    \caption{The experimental setup in which a Baxter robot has to insert a blue cylinder into a white tube (b). 
    The cylinder is with a string attached to the end-effector of the robot. 
    Camera images are recorded with the integrated end-effector camera.
    The sensor detecting the state of insertion is integrated into the white tube.
    Experiments on this platform were run fully autonomously without human intervention or simulations. }
    \label{Fig::baxter::setup}
\end{figure}

\section{Related Work}
The advancement of deep reinforcement learning in recent years has lead to the development of a number of methods combining model-free and model-based learning techniques, in particular to improve the sample complexity of deep reinforcement learning methods. 
Nagabandi et al. \cite{nagabandi2018neural} present a model-based deep reinforcement learning approach which learns a deep dynamic function mapping a state and action pair $(\klvec{s}_t, \klaction_t)$ to the next state $\klvec{s}_{t+1}$. 
The dynamics function is used to unroll a trajectory and to create an objective function based on the cumulative reward along the trajectory. 
This objective function is, then, used to optimize the actions along the trajectory and thereafter the first action is executed. 
The procedure is repeated whenever the next state is reached. 
After a dataset of executed trajectories is collected by the planning process, the policy of a model-free reinforcement learning algorithm is initialized in a supervised fashion by training it to match the actions produced by the planner. 
This technique is different to our approach in that we do not force the actor to match the executed action, but rather see it as an exploration from which we generate off-policy updates. 
Furthermore, it is implicitly assumed in~\cite{nagabandi2018neural} that a reward function is available for each state during the planning process. 
This can be a rather strong assumption, especially when learning in the real world without access to a simulation of the task and only providing minimal human supervision. 
Using executions in the environment during the planning process would be too costly since each change in state would require a re-execution of the whole trajectory. 
Since our insertion task provides only sparse rewards during execution, the trajectory planning algorithm would fail when relying only on rewards due to flat regions with zero reward and require additional reward engineering. 
This leaves a large and mostly flat region in the state space with a reward of zero. 

In \cite{chua2018deep}, Chua et al. introduce the model-based \textit{probabilistic ensembles with trajectory sampling} method. 
This work builds upon \cite{nagabandi2018neural}, but also makes use of a reward function. 
It makes use of a probabilistic formulation of the deep  dynamics function by using an ensemble of bootstrapped models encoding distributions to improve the sample complexity and improves the properties of the trajectory planner. 
Both approaches do not explicitly train an actor or a critic network. 

Similarly to us, Universal  planning networks \cite{srinivas2018universal} introduced by Srinivas et al. use a latent, gradient-based trajectory optimization method. 
However, the planner requires a goal state for the trajectory optimization.  
In certain tasks such as walking or running, it might be hard to acquire such a goal state to use in place of a velocity-based reward function. 
It is mentioned in \cite{srinivas2018universal} that to achieve walking, it was necessary to re-render images or reformulate the objective function by including an accessible dense reward function. 

In contrast to previous work, we focus explicitly on the impact of using trajectory optimization as an additional technique for exploration and its impact on the learning process when used by a deep reinforcement learning algorithm such as Deep Deterministic Policy Gradient. 
Furthermore, using an actor-critic architecture is a key element in our work to allow off-policy updates in a fast manner during the training process and to inform the trajectory optimization process initially.

\section{Method}
The following sections introduce the different components used to generate explorative actions via trajectory optimization. 
We first describe the image embedding used, then the training process of the dynamics function and Deep Deterministic Policy Gradient (DDPG) \cite{lillicrap2015continuous}, as well as its extension for the use of a value function. 
The section ends with a description of our trajectory optimization based exploration for DDPG.

\subsection{Image Embedding}
All tasks used throughout this paper are setup such that they use only images as observations, which have to be projected into a latent image embedding.
This serves two main purposes: First, the number of parameters is greatly reduced since the actor, critic, and the dynamics network can be trained directly in the low dimensional latent space. 
Second, it is desirable to enforce temporal constraints within the latent image embedding, namely that subsequent images are close to each other after being projected into the latent space. 
Therefore, we make use of the recently introduced approach of time-contrastive networks \cite{sermanet2017time}: 
the loss function enforces that the distance between latent representations of two subsequent images are small but the distance between two randomly chosen images is above a chosen threshold $\alpha$. 
Enforcing a temporal constraint in the latent space improves the learning process of a consistent deep dynamics function in the latent space~\cite{sermanet2017time}. 
Time-contrastive networks make use of two losses. 
The first is defined on the output of the decoder network and the input image as found in most autoencoder implementations. 
The second loss, the triplet loss, takes the latent representation $\mathbf{z}_t$ and  $\mathbf{z}_{t+1}$ of two temporally close images and the latent representation $\mathbf{z}_r$ of a randomly chosen image.

Thus, given two temporal images $\klim_t$ and $\klim_{t+1}$ and a randomly chosen image $\klim_{r}$, the loss functions for each element in the batch is given by
\begin{equation}
    L(\klim_t, \klim_{t+1}, \klim_r) = L_{\text{ae}}(\klim_t) + L_{\text{contr}}(\klim_t, \klim_{t+1}, \klim_r).
\end{equation}
The classical autoencoder loss $L_{\text{ae}}$ and the contrastive loss $L_{\text{contr}}$ are here defined as 
\begin{equation}
    \begin{split}
        &L_{\text{ae}} = \parallel \klim_t - \kldec(\klenc(\klim_t)) \parallel, \\
        &L_{\text{contr}} \left( \klim_t, \klim_{t+1}, \klim_r \right) = \parallel \klenc(\klim_{t}) - \klenc(\klim_{t+1})\parallel \\
         & \hspace{3ex}+ \max(\alpha - \parallel\klenc(\klim_{t}) - \klenc(\klim_r)\parallel, 0),
    \end{split}
\end{equation}
with $\klenc$ being the encoder and $\kldec$ being the decoder network.
The scalar value $\alpha$ defines the desired minimum distance between two random images in the latent embedding. 
Thus the classic autoencoder loss $L_{\text{ae}}$ trains both the encoder and decoder network to learn a reconstructable image embedding. 
The contrastive loss $L_{\text{contr}}$, on the other hand, generates only a learning signal for the encoder network and places a temporal constraint on the image embedding. 
The encoder and decoder consist of three convolutional networks with a kernel shape of $(3,3)$ and a stride of $(2,2)$, followed by a linear layer of size 20 and an l2-normalized embedding which projects the states on a unit sphere \cite{sermanet2017time}. All activation functions are rectified linear units (ReLU).

\subsection{Latent Dynamics}
Using a trajectory optimization algorithm in latent space requires a dynamics function which maps a latent state $\kllat_t$ and an action $\klaction_t$ to a subsequent latent state $\kllat_{t+1}$.
This allows us to unroll trajectories into the future. 
In the case of a single image with $\kllat_t=\klenc(\klim_t)$, we learn a dynamics mapping of $\kldyn(\kllat_t, \klaction_t)=\tilde{\kllat}_{t+1}$. 
In the other case, when our latent state is derived from several stacked images, then we project each image into the latent space, for example by
\begin{equation}
    \begin{bmatrix}
        \klenc(\klim_{t-2}) \\
        \klenc(\klim_{t-1}) \\
        \klenc(\klim_{t}) \\
    \end{bmatrix}
    =
    \begin{bmatrix}
        \klvec{z}_{t}^{t-2} \\
        \klvec{z}_{t}^{t-1} \\
        \klvec{z}_{t}^{t} \\
    \end{bmatrix}
    = \klvec{z}_t.
\end{equation}
To predict the next latent state, the dynamics function simply has to rotate the state and only predict the third latent sub-state. 
This function can be described with 
\begin{equation}
    \klvec{z}_t = 
    \begin{bmatrix}
        \klvec{z}_{t}^{t-2} \\
        \klvec{z}_{t}^{t-1} \\
        \klvec{z}_{t}^{t} \\
    \end{bmatrix}
    \mapsto 
    \begin{bmatrix}
        \klvec{z}_{t}^{t-1} \\
        \klvec{z}_{t}^{t} \\
        \overline{\kldyn}(\klvec{z}_t, \klvec{a}_t) \\
    \end{bmatrix}
    = \tilde{\klvec{z}}_{t+1},
\end{equation}
where $\overline{\kldyn}$ is the output of the neural network while we will use the notation $\kldyn(\kllat_t, \klaction_t)=\tilde{\kllat}_{t+1}$ for the whole operation, and $\tilde{\klvec{z}}_{i+1}$ is the predicted next latent state. 
The loss function for the dynamics network is then simply the difference between the predicted latent state and the actual latent state.
Therefore, the loss is given as
\begin{equation}
    \begin{split}
    L_{\text{dyn}}(\klim_{t-2:t}, \klaction_t, \klim_{t+1}) =~\parallel
    &\overline{\kldyn}(\kllat_{t}, \klaction_t) - \klenc(\klim_{t+1}) \parallel,
    \end{split}
\end{equation}
for each state-action-state triple $(\klim_{t-2:t}, \klaction_t, \klim_{t-1:t+1})$ observed during execution.
The dynamics networks is constructed out of 3 fully connected layers of size 400, 400 and 20 with ReLUs as nonlinear activation functions. 

\subsection{Deep Reinforcement Learning}
We make use of the Deep Deterministic Policy Gradient (DDPG) algorithm since action and state/latent space are continuous. 
DDPG is based on the actor-critic model which is characterized by the idea to generate a training signal for the actor (network) from the critic (network). 
In turn, the critic utilizes the actor to achieve an off-policy update and models usually a Q-value function. 
In DDPG, the actor is a network mapping (latent) states to an action with the goal of choosing optimal actions under a reward function.
Hence, the loss function for the actor is given by
\begin{equation}
    \begin{split}
        L_{\text{actor}}(\kllat_t) = -\klcriticQ(\kllat_t, \klact(\kllat_t)),
    \end{split}
\end{equation}
where only the parameters of the actor $\klact(\kllat_t)$ are optimized (see Eq. 6 in \cite{lillicrap2015continuous}). 
In the case of classical DDPG, the critic is a Q-function network, which maps state and action pairs to a Q-value: $Q(\kllat_t, \klaction_t) = r(\kllat_t, \klaction_t) + \gamma Q(\kllat_{t+1}, \klact(\kllat_{t+1}))$.
The scalar $gamma$ is a discount factor and $r(\kllat_t, \klaction_t)$ is the reward. 
The loss function of the critic network is based on the Bellman equation:
\begin{equation}
    \begin{split}
         L_{\text{critic}}(\kllat_t, \klaction_t, r_{t+1}, \kllat_{t+1}) = &\parallel \klcriticQ(\kllat_t, \klaction_t) - \\
        &~(r_{t+1} + \gamma \klcriticQ^\prime(\kllat_{t+1}, \klact^\prime(\kllat_{t+1}))) \parallel,
    \end{split}
\end{equation}
where $\klcriticQ^\prime$ and $\klact^\prime$ are target networks. 
For more details on DDPG we refer the interested reader to \cite{lillicrap2015continuous}.  
\begin{figure}
\centering
    \begin{subfigure}[b]{0.24\textwidth}
        \centering
        \includegraphics[height=4.5cm]{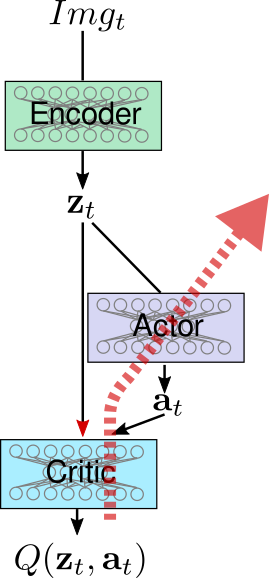}
        \caption{Q-Value based actor update}
    \end{subfigure}%
    \begin{subfigure}[b]{0.25\textwidth}
        \centering
        \includegraphics[width=0.95\textwidth]{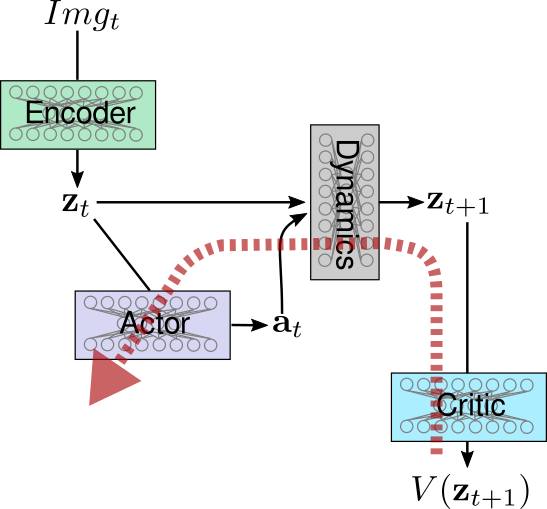}
        \caption{Value based actor update}
    \end{subfigure}
    \caption{The original DDPG algorithm (a) can be reformulated such that a value function (b) is used. 
    In the case of a value function the policy gradient (red arrow) is computed via a neural dynamics function.
    }
    \label{Fig::ddpg:q_and_v}
\end{figure}
It is worth noting that DDPG can be reformulated such that the critic resembles a value function instead of a Q-value function (Fig. \ref{Fig::ddpg:q_and_v}, see also \cite{heess2015learning}). 
A naive reformulation of the loss function given above is
\begin{equation}
    \begin{split}
        L_{\text{critic}}(\kllat_t, \klaction_t, r_t, \kllat_{t+1}) = &\parallel \klcriticV(\kllat_t) - (r_{t+1} + \gamma\klcriticV^\prime(\kllat_{t+1})) \parallel,
    \end{split}
\end{equation}
given an experience $(\kllat_t, \klaction_t, r_{t+1}, \kllat_{t+1})$. 
But this reformulation updates only on-policy and lacks the off-policy update ability of classical DDPG. 
Even worse, we would fail to use such a critic to update the actor since no action gradient can be computed due to the sole dependency on the state. 
However, since we have access to a dynamics function we reformulate for our extension of DDPG the loss function and incorporate off-policy updates with
\begin{equation}
    \begin{split}
        L_{\text{critic}}(\kllat_t, \klaction_t, r_t, \kllat_{t+1}) = &\parallel \klcriticV(\kllat_t)\\&- (r_t + \gamma \klcriticV^\prime(\kldyn({\kllat}_{t}, \klact^\prime(\kllat_t ))) \parallel.\\
    \end{split}
\end{equation}
This formulation allows for off-policy updates given the experience $(\kllat_t, \klaction_t, r_{t}, \kllat_{t+1})$, for which we assume that the reward $r(\kllat)$ is only state-dependent. 
While this might appear to be a strong assumption at first, it holds true for most tasks in robotics. 
The insertion task presented in the remainder of this paper is such a case in which the reward is fully described by the current position of both end-effector and the object to be inserted. 

The loss function for the actor is then given with
\begin{equation}
    \begin{split}
        L_{\text{actor}}(\kllat_t) = -\klcriticV(\kldyn(\kllat_t, \klact(\kllat_t))),
    \end{split}
\end{equation}
which is fully differentiable and, again, only used to optimize the parameters of the actor network.
We use for both actor and critic two fully connected hidden layers of size 400 and 300 with ReLUs as nonlinear activation functions. 

\subsection{Optimized Exploration}
\begin{figure}
    \centering
    \includegraphics[width=0.45\textwidth]{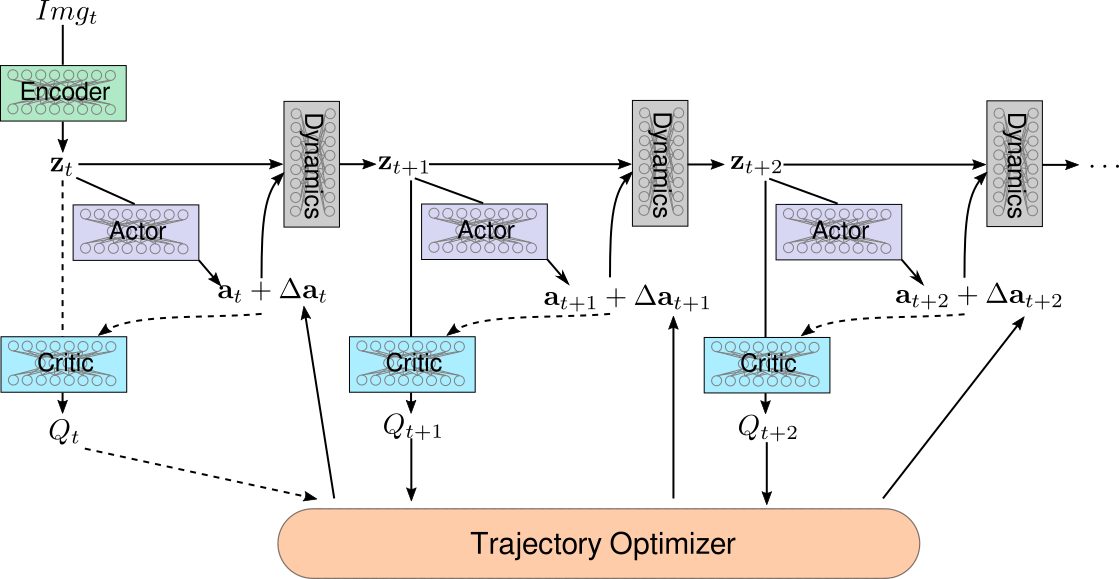}
    \caption{The proposed exploration strategy unrolls the trajectory in the latent space and uses the Value/Q-Value to optimize the actions of the trajectory. 
    Dotted connections might not be used when using a Value function as critic. }
    \label{Fig::traj_opt}
\end{figure}
Due to the deterministic nature of the actor network in DDPG and similar algorithms, the standard approach for exploration is to add random noise to actions. 
Random noise is usually generated from an Ornstein-Uhlenbeck process or a Gaussian distribution with fixed parameters. 
Such parameters, like the variance for a Gaussian distribution, are usually chosen by intuition or have to be optimized as hyper-parameter, for example with grid-search. 
In preliminary experiments we found Ornstein-Uhlenbeck processes with $\sigma=0.5$ and $\theta=0.15$ most effective on the chosen simulated task. 
In the presented approach we make use of the fact that we can access a dynamics function and therefore unroll trajectories throughout the latent space. 
The basic idea is to first unroll a trajectory using the actor network a number of steps into the future from the current point in time. 
We then optimize the actions $\klaction_t, \cdots, \klaction_{t+n}$ such that we maximize the Q-values/rewards along the latent trajectory. 
We characterize a latent trajectory, given a start state $\kllat_t = \klenc(\klim_t)$, as a sequence of state-action pairs $(\kllat_t, \klaction_t, \cdots, \kllat_{t+H}, \klaction_{t+H}, \kllat_{t+H+1})$. 
We can then formulate a scalar function to be maximized by the trajectory optimizer based on the Q-value or reward-functions available. 
This process is visualized in Fig. \ref{Fig::traj_opt}. 
The Q-function in the following equations can be substituted with a learned value function. 
An intuitive objective function to optimize is to simply sum up all Q-values for each state-action pair of the trajectory
\begin{equation}
    \begin{split}
        f_{Q}(\klvec{a}_{t:t+H}, \kllat_t) = w_{0} Q(\kllat_{t}, \klaction_t) + \sum_{j=1}^{H} w_{j} Q(\kllat_{t+j}, \klaction_{t+j}),
    \end{split}
    \label{Eq::Q}
\end{equation}
with $\kllat_{t+j} = \kldyn(\kllat_{t+j-1}, \klaction_{t+j-1})$ and $\kllat_t$ being the current state from which we start unrolling the trajectory. 
The time-dependent weight $w_i$ determines how much actions are going to be impacted by future states and their values and can be uniform, linearly increasing or exponential.
We consider in our experiments the special case of $w_i=\frac{1}{H}$. 
Alternatively, if one has access to a rewards function or learns a state-to-reward mapping simultaneously, then an objective function can be used which accumulates all rewards along the latent trajectory and adds only the final q-value:
\begin{equation}
    \begin{split}
     f_{r+Q}(\klvec{a}_{t:t+H}, \klvec{z}_t) =& \sum_{j=1}^{H-1} w_j r(\kllat_{t+j})  + w_{H}Q(\kllat_{t+H}, \klaction_{t+H}).
    \end{split}
    \label{Eq::QR}
\end{equation}
Clearly, this objective function is especially useful in the context of tasks with dense rewards. 
Both objective functions will be evaluated on the simulated cheetah task, which provides such dense rewards. 
While executing policies in the real world, we unroll a planning trajectory from the current state for $n$ steps into the future. 
Then, the actions $\klaction_{t:t+H}$ are optimized under one of the introduced objectives from above with a gradient-based optimization method such as L-BFGS \cite{zhu1997algorithm}. 
After a number of iterations of trajectory optimization, here 20, the first action of the trajectory, namely $\klaction_t$, is executed in the real world (Alg. \ref{alg1}).

\begin{algorithm}
\caption{Exploration through trajectory optimization in DDPG} 
\label{alg1} 
\begin{algorithmic} 
    \REQUIRE Horizon $H$, Encoder network
    \FOR{number of episodes}
        \WHILE{end of episode not reached}
            \STATE {Compute latent state $\kllat_t$ from images with encoder}
            \STATE{Initialize action with $\klaction_t = \klact(\kllat_t)$}
            \IF{ training}
                \FOR{ $k=t+1:t+H$}
                    \STATE{Initialize action with $\klaction_k = \klact(\kllat_k)$}
                    \STATE{Predict latent state $\kllat_{k+1} = \kldyn(\kllat_k, \klaction_k)$}
                \ENDFOR
                \STATE{Optimize $\max_{\klvec{a}_{t:t+H}} f(\klvec{a}_{t:t+H}, \klvec{z}_t)$ }
            \ENDIF
            \STATE{Execute step in environment with action $\klvec{a}_t$}
            \STATE{Store $(\kllat_t, \klaction_t, r_t, \kllat_{t+1})$ in replay buffer}
        \ENDWHILE
            \STATE{Optimize dynamics network} 
            \STATE{Optimize actor network}
            \STATE{Optimize critic network}
            \STATE{Update target networks}
    \ENDFOR
\end{algorithmic}
\end{algorithm}

\section{Experiments}
We compare in our experiments the classical approach of exploration in DDPG with an optimized Ornstein-Uhlenbeck process against the introduced approach of exploration through optimization. 
First, an experiment in simulation was conducted using the DeepMind Control Suite \cite{tassa2018deepmind}. 
The cheetah task, in which a two-dimensional bipedal agent has to learn to walk, is especially interesting because it involves contacts with the environment that makes the dynamics hard to model. 
In the second experiment, we evaluate the algorithms directly on a robot and aim to solve an insertion task in the real world.

\subsection{Evaluation in Simulation on the Cheetah Task}
\begin{figure*}
    \centering
    \begin{subfigure}[b]{0.4\textwidth}
        \includegraphics[width=\textwidth]{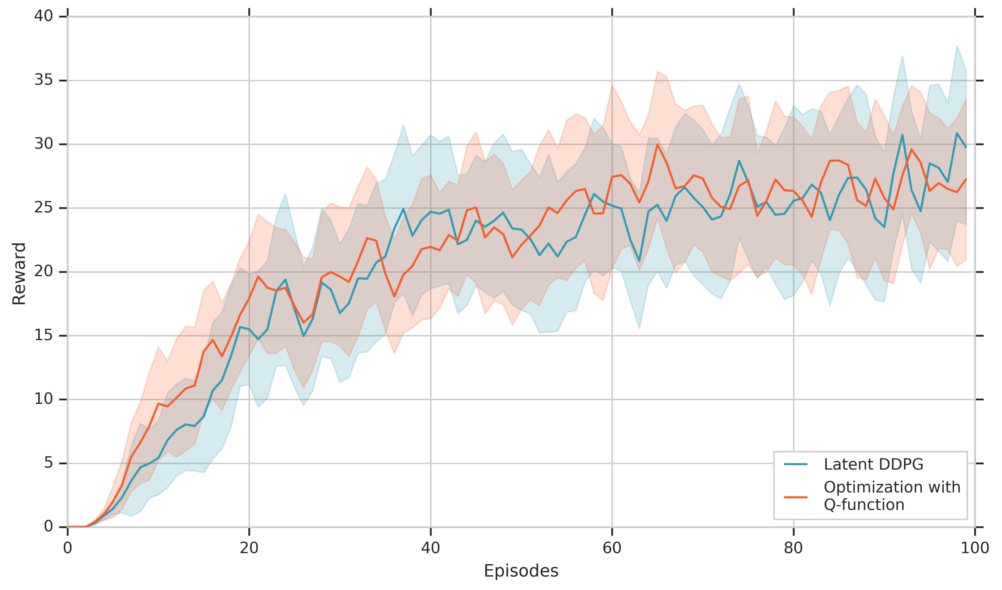}
        \caption{Deterministic Policy (Q-Value)}
    \end{subfigure}%
    \begin{subfigure}[b]{0.4\textwidth}
        \includegraphics[width=\textwidth]{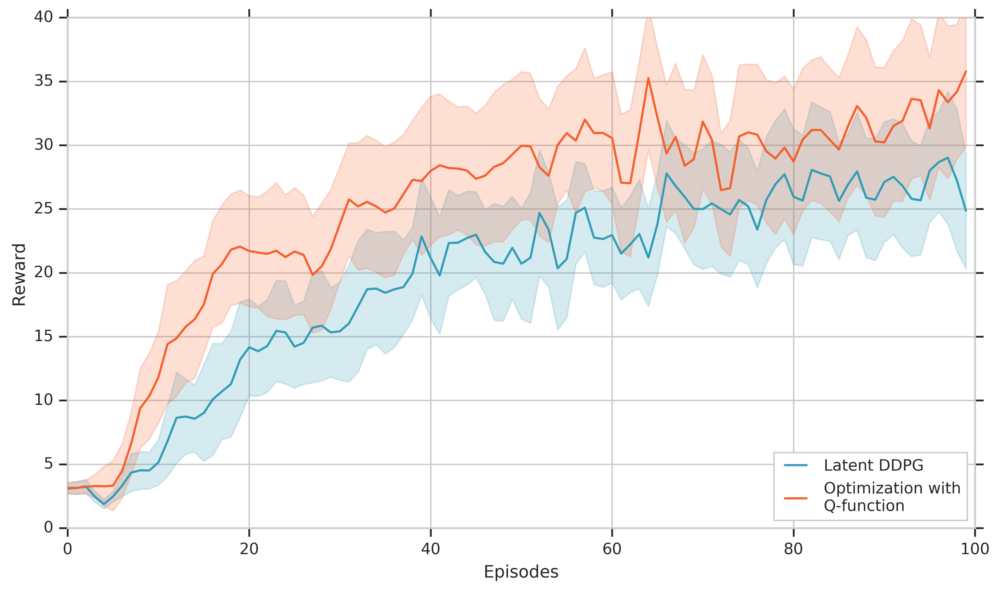}
        \caption{Exploration (Q-Value)}
    \end{subfigure}
    \begin{subfigure}[b]{0.4\textwidth}
        \includegraphics[width=\textwidth]{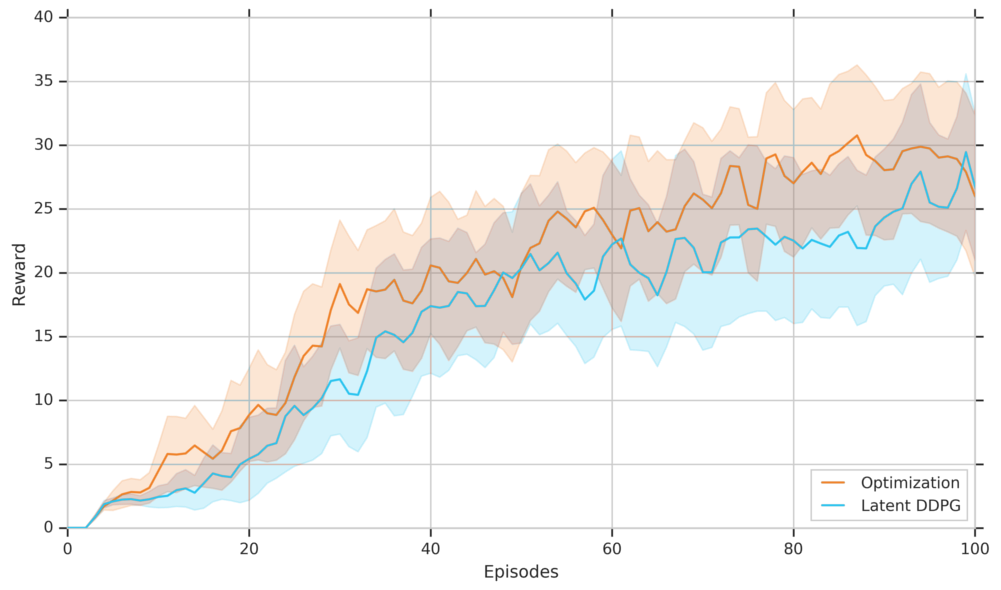}
        \caption{Deterministic Policy (Value)}
    \end{subfigure}%
    \begin{subfigure}[b]{0.4\textwidth}
        \includegraphics[width=\textwidth]{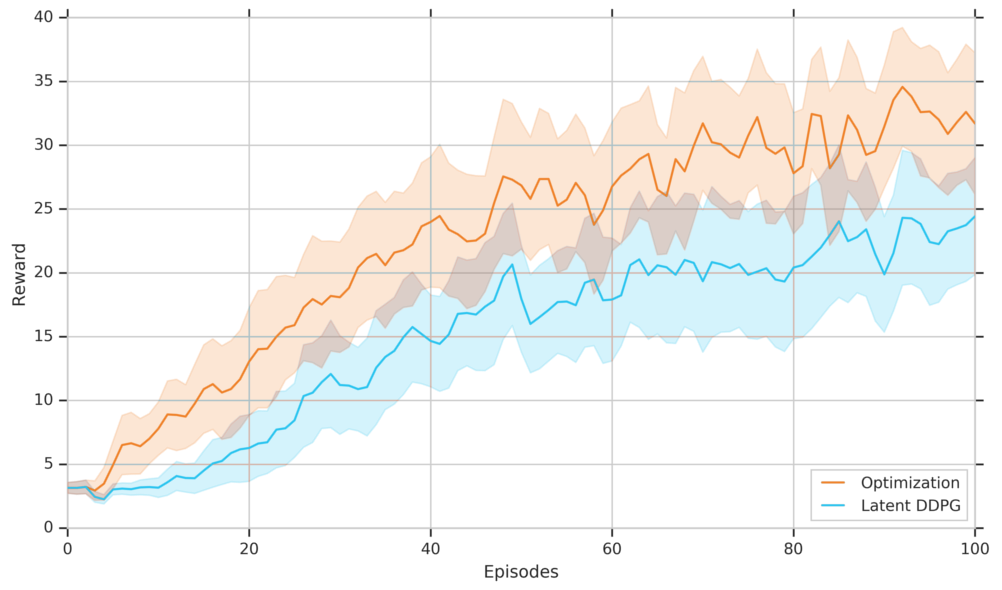}
        \caption{Exploration (Value)}
    \end{subfigure}
    \caption{Comparison between DDPG using exploration with optimization (orange) and classical exploration using an Ornstein-Uhlenbeck process (blue) on the simulated cheetah task. The exploitation graph shows the evaluation of actions produced by the deterministic actor while exploration strategies are applied during training. 
    }
    \label{Fig::cheetah::basic}
\end{figure*}

\begin{figure}
    \centering
    \includegraphics[width=0.4\textwidth]{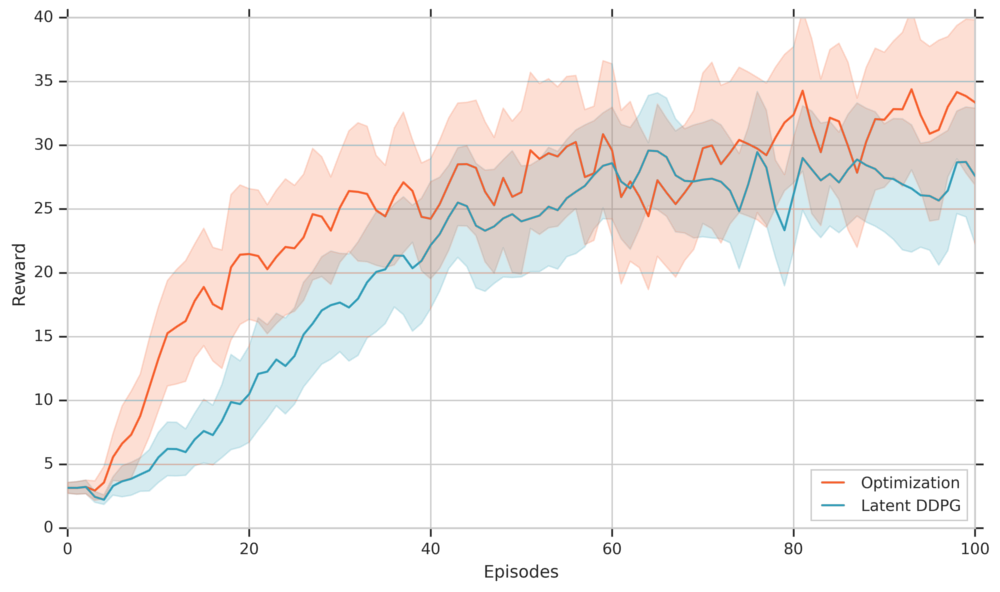}
    \caption{Comparison between DDPG using exploration with optimization (orange) and classical exploration using an Ornstein-Uhlenbeck process (blue) on the simulated cheetah task while using a value function as critic. 
    The number of training iterations per episode were raised from 1000 (Fig. \ref{Fig::cheetah::basic}-d) to 3000 for this evaluation.
    }
    \label{Fig::cheetah:3000}
\end{figure}

The cheetah environment of the DeepMind control suite \cite{tassa2018deepmind} has six degrees-of-freedom in its joints and we only use camera images as state information. 
The actions are limited to the range of $[-1,1]$ and camera images are of the size $320\times240~px$ in RGB and were resized to $64\times64~px$. 
Each episode consists of 420 time steps and actions are repeated two times per time step. 
First, a dataset of 50 representative episodes was collected through the use of DDPG on the original state space of joint positions, joint velocities, relative body pose and body velocity of cheetah. 
This dataset was used to train the time-contrastive autoencoder as described above. 
The same parameters for the neural encoder were use for all exploration strategies. 
This was done to allow the sole evaluation of the exploration strategies independently of the used embedding. 
Since cheetah is a quite dynamic task and rewards depend on the forward velocity, this velocity must be inferable from each state.
Hence, we project three subsequent images $(\klim_{t-2}, \klim_{t-1}, \klim_t)$ down by using the encoder network and define the current state $\kllat_t$ as the three stacked latent states $\kllat_t=[\kllat_{t-2}, \kllat_{t-1}, \kllat_{t}]^T$. 
For each of the presented evaluations 25 experiments were executed and the mean and standard deviations of the episodic cumulative rewards are shown in Figures \ref{Fig::cheetah::basic}-\ref{Fig::objectives}. 


\subsubsection{Comparison between Ornstein-Uhlenbeck and optimized exploration}
As a first step we optimized the hyperparameter $\sigma$ of DDPG and found that an Ornstein-Uhlenbeck process with $\sigma=0.5$ and $\theta=0.15$ achieve a better result for DDPG on this task than the variance of $\sigma=0.2$ proposed in \cite{lillicrap2015continuous}, especially in the early stages of the training process. 
A planning horizon of ten steps was used to generate the optimized noise. 
We make comparisons between the training process, in which we use the exploration strategies, and the test case, in which we execute the deterministic actions produced by the actor without noise. 
Throughout the training process we evaluate the current policy of the actor after each episode. 
The results are presented in Fig. \ref{Fig::cheetah::basic}. 

\subsubsection{Comparison between different planning horizons}
\begin{figure}
    \centering
    \begin{subfigure}[b]{0.4\textwidth}
        \includegraphics[width=\textwidth]{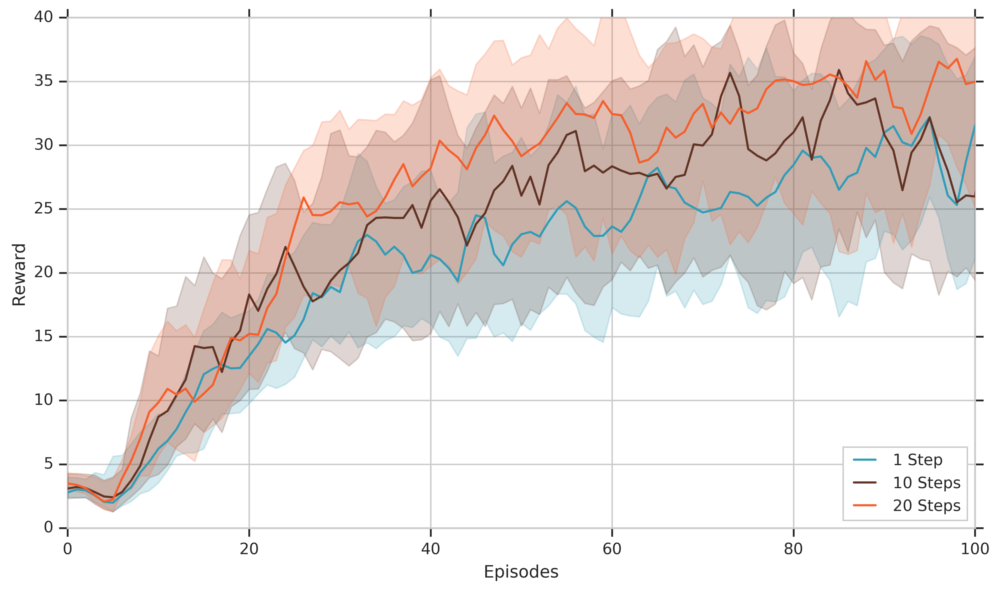}
    \end{subfigure}
    \caption{Exploration through optimization evaluated with different horizons for the planning trajectory on the simulated cheetah task. 
     }
    \label{Fig::cheetah::steps}
\end{figure}

The main hyperparameter for optimized noise is the length of the planning horizon. 
If it is too short, actions are optimized greedily for immediate or apparent short-term success; if it is too long, the planning error becomes too large. 
Figure \ref{Fig::cheetah::steps} shows the optimized exploration strategy with three different step-sizes: one step, ten steps and 20 steps into the future from the current state. 

\subsubsection{Comparison between different objectives}
\begin{figure}
    \centering
    \includegraphics[width=0.4\textwidth]{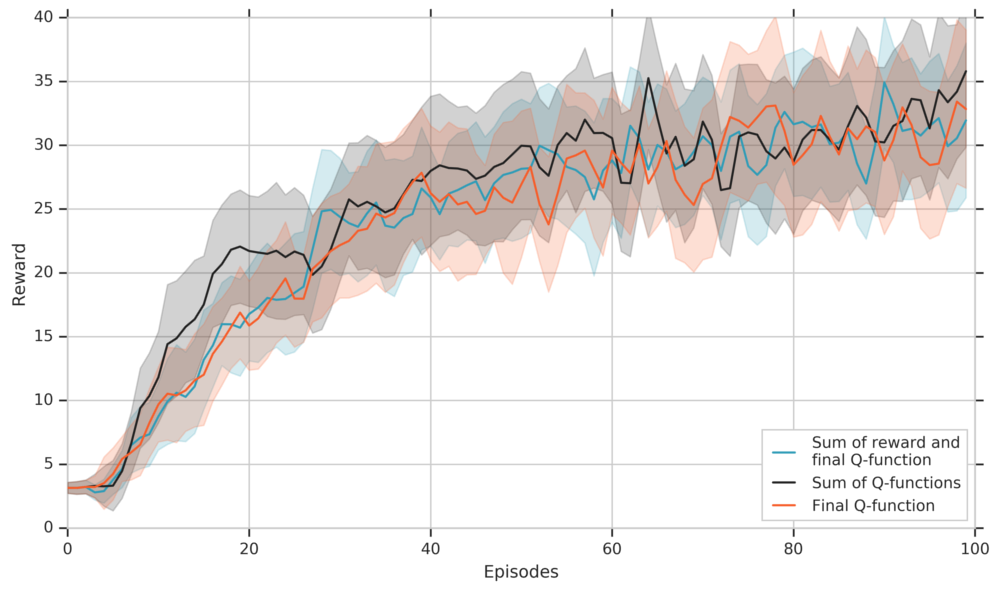}
    \caption{Comparison between three different objective functions for optimized exploration on the simulated cheetah task. 
     }
    \label{Fig::objectives}
\end{figure}

We introduced two potential objective functions, based on Q-values (Eq. \ref{Eq::Q}) and a mix of reward- and Q-function (Eq. \ref{Eq::QR}). 
We compare both of these against another objective where we only optimize for the q-value of the very last state-action pair of the unrolled trajectory (Fig. \ref{Fig::objectives}). 

\captionsetup{belowskip=0pt}
\begin{figure}
    \centering
    \begin{subfigure}[b]{0.5\textwidth}
    \includegraphics[width=\textwidth]{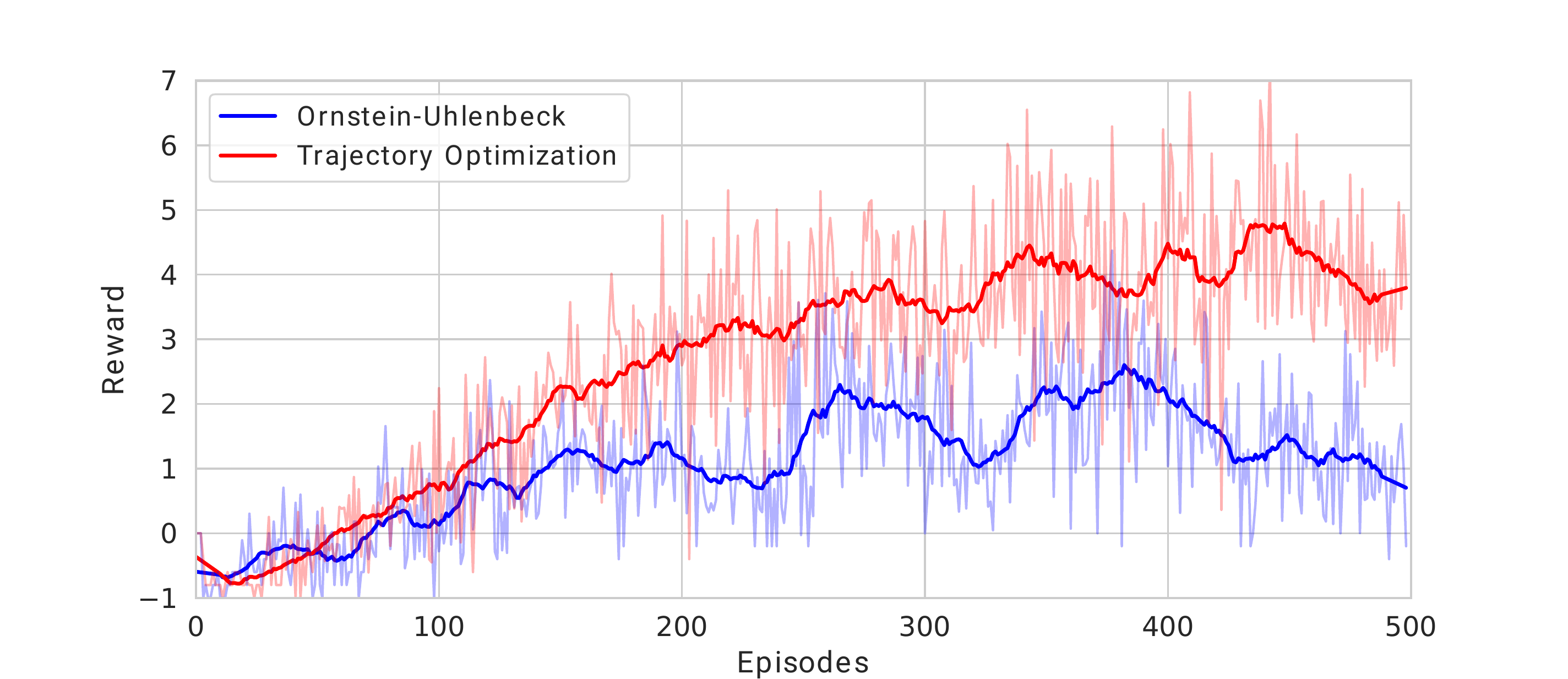}
        \caption{Deterministic Policy}
    \end{subfigure}
    \begin{subfigure}[b]{0.5\textwidth}
        \includegraphics[width=\textwidth]{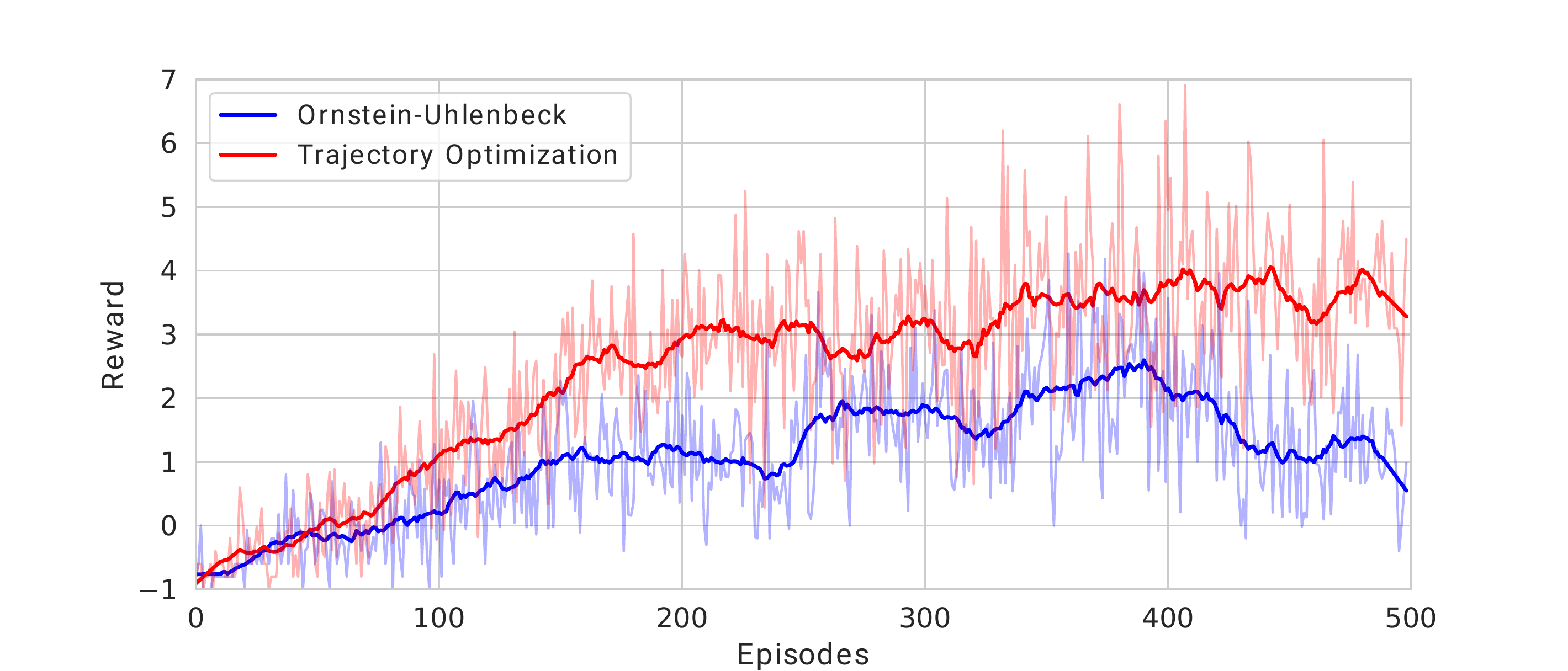}
        \caption{Exploration}
    \end{subfigure}
    \caption{Comparison between exploration with an Ornstein-Uhlenbeck (blue) and exploration through optimization (red) on the insertion task in the real world. 
    The planning horizon is three steps. 
    The figures show the cumulative rewards averaged over five experiments in light colours and in bold colours, for better interpretability due to the sparse reward, the mean smoothed with a Savitzky-Golay filter with window size 21 and 1st order polynomials. 
    }
    \label{Fig::baxter::basic}
\end{figure}

\begin{table}
\caption{The average success rate of insertion for policies trained by DDPG with standard Ornstein-Uhlenbeck exploration or trajectory optimization with varying planning horizons. The individual success rates for each experiment were computed over a window of 50 subsequent episodes of 500 executions total. The average success rates and standard deviations were then computed with the highest success rate achieved in each experiment. A total of five experiments were executed for each method. }
\label{Tabel::Horizons}
\centering
\begin{tabular}{l|c}
Method             & Avg. Success rate $(\pm std)$         \\ \hline\hline
Ornstein-Uhlenbeck Exploration & $75.2\% ~(\pm 11.7\%)$ \\
1 Step Planning Horizon             & $\mathbf{93.2}\% ~(\pm 5.2\%)$  \\
3 Steps Planning Horizon            & $91.6\% ~(\pm \mathbf{1.5}\%)$  \\
5 Steps  Planning Horizon           & $84.0\% ~(\pm 14.1\%)$   \\
15 Steps Planning Horizon           & $84.4\% ~(\pm 9\%)$   
\end{tabular}
\end{table}
\subsection{Insertion in the real world}
Fast exploration is especially important when tasks have to be solved in a real world environment and training needs to be executed on the real robot. 
An insertion task was set up in which a Baxter robot had to insert a cylinder into a tube where both training and testing were performed in the real world environment, without the use of simulation (Fig. \ref{Fig::baxter::setup}) \footnote{A video of the experiment can be found here: \url{https://youtu.be/rfZcUWnut5I}}.
Cylinder and tube were 3D-printed.
The cylinder was attached to the right end-effector of the robot with a string. 
The position control mode was used because there is a variable delay in the observations. 
Image observation were acquired from the end-effector camera of the Baxter robot via ethernet. 
The six dimensional actions are in the range of $[-0.05, 0.05]$ radians and represent the deviation for each joint of the arm at a point in time.  
This restriction ensures a strong correlation between subsequent camera images throughout the execution and allows the task to be solved in 20 steps. 
The initial position (radians) of the robot arm was randomized by sampling from a normal distribution with mean $\mu_{1:6}=(0.48, -1.23, -0.15,  1.42,  0.025, 1.35)$ and variances $\sigma_{1:6}=(0.05, 0.05, 0.05, 0.05, 0.05, 0.1)$, ensuring that the tube is in the image. 
As a simplification of the task, we excluded the last rotational wrist joint of the robot arm. 
Because of the adynamic nature of this task and the necessity to use position control mode it is sufficient to use the latent representation of the current image versus a stack of images as in simulation. 
Larger movements of the cylinder appear as blur in the images. 
Each episode consists of 20 time steps and a sparse reward is used: 
For safety reasons, if the end-effector left the designated workspace area, the episode ended and a reward of $-1$ is assigned. 
When the cylinder is inserted into the tube, the extent of insertion is transformed into a reward from $[0, 1.0]$ and an episode stops if a reward of $0.9$ or higher is assigned. 
The state of insertion is measured with a laser-based time-of-flight sensor (VL6180). 
The reward for all other possible states is zero. 
Five experiments were conducted on the robot: DDPG with a value function as critic and Ornstein-Uhlenbeck exploration, DDPG with exploration using trajectory optimization and a varying planning horizon (1, 3, 5 and 15 steps). 
We use a reduced planning horizon in this task due to the low number of time steps per episode. 
The comparison between Ornstein-Uhlenbeck exploration and optimized exploration with a horizon of three is shown in Fig. \ref{Fig::baxter::basic}. 
Every episode which ends with a negative cumulative reward violated the workspace boundaries and episodes reaching a reward of 0.9 or more were successful insertions. 
Table \ref{Tabel::Horizons} shows the comparison between exploration with Ornstein-Uhlenbeck noise and using planning horizons of different lengths in terms of successful insertions. 
Each experiment was repeated five times and the cumulative reward for each episode is used to compute the mean shown in Figure \ref{Fig::baxter::basic}. 
For better interpretability, the figures show, in bold lines, additionally a smoothed version of the mean where a Savitzky-Golay filter was applied with a window size of 21 and polynomials of order one. 
The autoencoder network as well as the dynamics network were trained with a demonstration dataset of 50 trajectories. 
Of these, 19 were positive demonstrations, in which the cylinder was successfully inserted. 
At the beginning of each training process, 5 of these 19 trajectories were added to the replay buffer to ensure convergence of the training process due to the difficulty of the task caused by using sparse reward.

\section{Discussion}
We start with a discussion of the results from the simulated bipedal cheetah task which uses a dense reward function:
The first insight is that both actors seem to perform equally well after 20 episodes, with the actor trained with optimized noise outperforming classical DDPG throughout the first 20 episodes (Fig. \ref{Fig::cheetah::basic} (a)). 
However, during training the optimized exploration does not only perform better than exploration with an Ornstein-Uhlenbeck process (Fig. \ref{Fig::cheetah::basic} (b)) but also performs better than the actions produced by both actors during test time (Fig. \ref{Fig::cheetah::basic} (a)). 

We found that using a critic network modelling the Q-function (Fig. \ref{Fig::cheetah::basic} (b)) outperformed the formulation of DDPG using a value network when using optimized exploration (Fig. \ref{Fig::cheetah::basic} (d)), while DDPG with Ornstein-Uhlenbeck noise performs slightly better with a Value network (Fig. \ref{Fig::cheetah::basic} (a,d)). 
One could argue, that the effects of using optimized noise could vanish when increasing the number of trainings per episode, giving DDPG more time to find an optimal actor given the current training set. 
Following this line of thought we increased the number of training iterations per episode three times to 3000 (Fig. \ref{Fig::cheetah:3000}). 
The evaluation shows that while DDPG with OU noise improves in the later stages of the learning process, the trajectory optimization  uncovers valuable training experience now much faster early on. 
This strongly indicates that the data distribution generated by the exploration strategy has an impact on the performance of DDPG. 
Evaluating the step-lengths we could find that trajectory optimization improved up to a planning horizon of 20 steps, although we opted for our experiments with a conservative planning horizon of 10 steps to reduce the overall training time.  
The evaluation of the three introduced objective functions show that the summation of Q-values along the planning trajectory yields better performance in the early training stages, up to episode 25, for the dense reward task (Fig. \ref{Fig::objectives}). 
This is an interesting result given that many other trajectory optimization approaches use a Bellman-inspired sum of weighted rewards \cite{nagabandi2018neural,chua2018deep}. 
It is also worth to notice that the Q-Value is the more suitable objective function for optimizing actions in the presented real-world insertion task due to the reward function being zero for the majority of time steps. 

The results showing the learning progress on the insertion task in the real world draw a clearer picture of the benefit of exploration through optimization (Fig. \ref{Fig::baxter::basic}, Table \ref{Tabel::Horizons}).  
Generally, after roughly 50 training episodes, the networks trained with optimized exploration outperformed DDPG with OU and also achieved higher rewards in later stages of the learning process (Fig. \ref{Fig::baxter::basic}). 
An evaluation of the length of the planning horizon shows, as expected, that longer planning horizons lead to a decreases performance (Table \ref{Tabel::Horizons}). 
This is very likely due to the accumulating error of predicted future states from the dynamics network. 
However, even with longer planning horizons the presented approach outperformed exploration using OU noise. 

\section{Conclusion}
This work investigated the possibility of combining an actor-critic reinforcement learning method with a model-based trajectory optimization method for exploration. 
By using trajectory optimization only to gain new experience, the ability of DDPG to learn an optimal policy is not affected and we can furthermore make use of DDPG's off-policy training ability. 
We were able to show that by using this strategy, a performance gain can be achieved, especially in the presented real world insertion task learned from images. 
It is worth noting that this performance gain can be mainly attributed to the change in exploration strategy since a fixed image embedding was used, reducing the possibility of performance differences caused by using different image embeddings. 
This work only considered using reward, Q-Value or value functions as objective functions for optimizing the latent trajectory. 
In future work we plan to investigate the possibility of using additional cost terms, eg. safety and state-novelty. 
Furthermore, another natural next step would be to use probabilistic dynamics networks and advanced trajectory optimization algorithms to evaluate their impact on deep reinforcement learning algorithms when used for exploration in this setup.

\bibliographystyle{plain}
\bibliography{bib}

\end{document}